# Learning to Order BDD Variables in Verification

**Orna Grumberg**                                          ORNA@CS.TECHNION.AC.IL
**Shlomi Livne**                                           SLIVNE@CS.TECHNION.AC.IL
**Shaul Markovitch**                                       SHAULM@CS.TECHNION.AC.IL
*Computer Science Department*
*Technion - Israel Institute of Technology*
*Haifa 32000, Israel*

## Abstract

The size and complexity of software and hardware systems have significantly increased in the past years. As a result, it is harder to guarantee their correct behavior. One of the most successful methods for automated verification of finite-state systems is *model checking*. Most of the current model-checking systems use binary decision diagrams (BDDs) for the representation of the tested model and in the verification process of its properties. Generally, BDDs allow a canonical compact representation of a boolean function (given an order of its variables). The more compact the BDD is, the better performance one gets from the verifier. However, finding an optimal order for a BDD is an NP-complete problem. Therefore, several heuristic methods based on expert knowledge have been developed for variable ordering.

We propose an alternative approach in which the variable ordering algorithm gains "ordering experience" from training models and uses the learned knowledge for finding good orders. Our methodology is based on offline learning of pair precedence classifiers from training models, that is, learning which variable pair permutation is more likely to lead to a good order. For each training model, a number of training sequences are evaluated. Every training model variable pair permutation is then tagged based on its performance on the evaluated orders. The tagged permutations are then passed through a feature extractor and are given as examples to a classifier creation algorithm. Given a model for which an order is requested, the ordering algorithm consults each precedence classifier and constructs a pair precedence table which is used to create the order.

Our algorithm was integrated with SMV, which is one of the most widely used verification systems. Preliminary empirical evaluation of our methodology, using real benchmark models, shows performance that is better than random ordering and is competitive with existing algorithms that use expert knowledge. We believe that in sub-domains of models (alu, caches, etc.) our system will prove even more valuable. This is because it features the ability to learn sub-domain knowledge, something that no other ordering algorithm does.

## 1. Introduction

The size and complexity of software and hardware systems have significantly increased in the past years. As a result, it is harder to guarantee their correct behavior. Thus, formal methods, preferably computerized, are needed for this task.

One of the most successful methods for automated verification of finite-state systems is *temporal logic model checking* (Clarke, Emerson, & Sistla, 1986; Queille & Sifakis, 1981). Temporal logics are suitable formalisms for describing the behavior of a program over time. A model checking procedure receives a finite-state model of the system and a specification written as a temporal logic formula. It returns "yes" if the model satisfies the formula (meaning that the system behaves according to the specification). Otherwise, it returns "no", along with a counter example that demonstrates a bad behavior.

Model checking has been very successful in finding subtle errors in various systems. It is currently recognized by the hardware industry as an important component of the development phase of new designs. However, model checking procedures often suffer from high space requirements, needed for holding the transition relation and the intermediate results.

One of the most promising solutions to this problem is the use of *binary decision diagrams* (BDDs) (Akers, 1978; Bryant, 1986) as the basic data structure in model checking. BDDs are canonical representations of boolean functions and are often very concise in size. Their conciseness also yields efficiency in computation time. Since it is straightforward to represent the transition relation and the intermediate results as boolean functions, BDDs are particularly suitable for model checking. Today, existing industrial BDD-based verifiers, such as IBM's RuleBase (Beer, Ben-David, Eisner, & Landver, 1996) and Motorola's Verdict (Kaufmann & Pixley, 1997) are used by many companies in their development infrastructure.

The size of a BDD for a given function is sensitive to the ordering of the variables in the BDD. However, finding an optimal ordering, which yields a smallest BDD for a given function, is an NP-complete problem (Bollig & Wegener, 1996). Therefore, several heuristic algorithms based on expert knowledge have been developed for variable ordering in the hope of reducing the BDD size. Unfortunately, and in spite of the resources invested, these algorithms do not produce good enough variable orders. The reason for this may be that only general rules are used and no domain-specific knowledge is exploited.

The goal of this research is to develop learning techniques for acquiring and using domain-specific knowledge for variable ordering. We assume the availability of one or more training models. The training models are used for off-line acquisition of ordering experience which can be used for ordering variables of a previously unseen model.

We first present a method for converting the ordering learning task into a concept learning problem. The concept is the set of all ordered variable pairs that are in the "right" order. The examples are ordered pairs of variables of a given training model. We show a statistical method for tagging examples based on evaluated training orders and present a set of variable-pair features. The result is a standard concept learning problem. We apply decision tree learning to generate a decision tree for each training model. When used for an unseen model, we combine the trees and generate a partial order which is used for generating the required order. We also present an extension of the algorithm which learns context-based precedence relations.

Our algorithm was integrated with SMV (McMillan, 1993), which is the backbone of many verification systems. Empirical evaluation of our methodology, using real benchmark models of hardware designs, shows performance that is much better than random ordering and is competitive with existing algorithms that use expert knowledge.

Section 2 contains background on model checking. Section 3 presents our main algorithm with empirical evaluation. Section 4 shows the context-based algorithm. Our conclusions are presented in Section 5.

## 2. Background

*Model checking* was introduced by Clarke and Emerson (1986) and by Queille and Sifakis (1981) in the early 1980s. They presented algorithms that automatically reason about temporal properties of finite state systems by exploring the state space. The use of *binary decision diagrams* (BDDs) to represent finite state systems and to perform symbolic state traversal is called *symbolic model checking*. The use of BDDs has greatly extended the capacity of model checkers. Models with $2^{100}$ and more states are routinely being verified.

BDDs were introduced by Akers (1978) as compact representations for boolean functions. Bryant (1986) proposed *ordered binary decision diagrams* (OBDDs) as canonical representations of boolean functions. He also showed algorithms for computing boolean operations efficiently on OBDDs.

The following subsection gives an overview of how finite state systems are represented in symbolic model checking. BDDs are then described and the variable ordering problem is defined. Existing algorithms for static variable ordering algorithms are reviewed. Finally, a brief description of machine learning algorithms used for ordering is given.

### 2.1 Finite State Machines in Symbolic Model Checking

Finite state systems (FSM) can be described by defining the set of possible states in a system and the transition relation between these states. A state typically describes values of components (e.g., latches in digital circuits), where each component is represented by a *state variable*. Let $V = \{v_0, v_1, ...v_{n-1}\}$ be the set of variables in a system. Let $K_{v_i}$ be the set of possible values for variable $v_i$. Then a state in the system can be described by assigning values to all the variables in $V$. The set of all possible states $S_A$ is

$$S_A = K_{v_0} \times K_{v_1} .... \times K_{v_{n-1}}.$$

A state can be written using a function that is true only in this state:

$$\bigwedge_{i=0}^{n-1} (v_i == c_j),$$

where $c_j \in K_{v_i}$ is the value of $v_i$ in the state. A set of states can be described by a function as the disjunction of the functions that represent the states.

Figure 1 shows a 3-bit counter. A state in the 3-bit counter can be described by a tuple which gives an assignment to the 3 variables $v_2, v_1, v_0$. For example, the tuple $\langle 1, 0, 0 \rangle$ represents the state with $v_2 = 1$, $v_1 = 0$, $v_0 = 0$. The corresponding boolean expression for the state is $(v_2 == 1) \wedge (v_1 == 0) \wedge (v_0 == 0)$.

In order to describe a system, we also need to specify its transition relation. The transition relation describes all the possible transitions of each system state. It can thus be described by pairs of states, $\langle present\ state,\ next\ state \rangle$, where next state is a system state after a transition from the present state. The variables in $V$ will represent the *present state variables*, and for each variable $v_i \in V$ we will define a corresponding *next state variable* $v_i' \in V'$. $V'$ will denote the set of next state variables.

An example of a valid transition for the 3-bit counter is from $\langle 0, 0, 0 \rangle$ to $\langle 0, 0, 1 \rangle$. The boolean function which represents this transition is $(v_2 == 0) \wedge (v_1 == 0) \wedge (v_0 == 0) \wedge (v_2' == 0) \wedge (v_1' == 0) \wedge (v_0' == 1)$. The transition relation can be represented by a boolean function which is the disjunction of the boolean functions of each of the transitions. Table 1 shows the transition relation for the 3-bit counter.

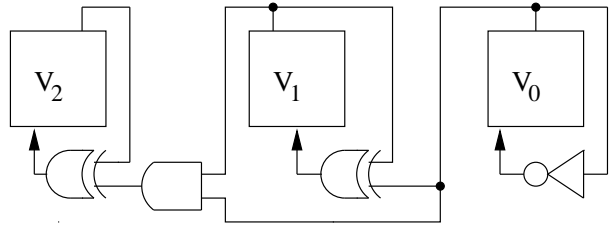

Figure 1: 3 bit counter

| Present State | | | Next State | | |
|---|---|---|---|---|---|
| $v_2$ | $v_1$ | $v_0$ | $v_2'$ | $v_1'$ | $v_0'$ |
| 0 | 0 | 0 | 0 | 0 | 1 |
| 0 | 0 | 1 | 0 | 1 | 0 |
| 0 | 1 | 0 | 0 | 1 | 1 |
| 0 | 1 | 1 | 1 | 0 | 0 |
| 1 | 0 | 0 | 1 | 0 | 1 |
| 1 | 0 | 1 | 1 | 1 | 0 |
| 1 | 1 | 0 | 1 | 1 | 1 |
| 1 | 1 | 1 | 0 | 0 | 0 |

Table 1: 3-bit counter transition relation table

An alternative method for describing the transition relation is for each state variable to define its valid next states. This form is known as the *partitioned transition relation*. The transition relation is then described by a set of functions (instead of one), one for each variable. For variable $v_i$, a boolean function $T_i(V, v_i')$ defines the next value of $v_i$, $v_i'$, given that the current state of the system is $V$.

For synchronous systems, in which there is a simultaneous transition of all the system components, the transition relation is

$$\bigwedge_{i=0}^{n-1} T_i(V, v_i').$$

In model checking it is common to use the partitioned transition relation form of representation, since it is usually more compact in memory requirements and thus allows the handling of larger systems. For the 3-bit counter, the next state boolean functions are given below, where $\otimes$ stands for the boolean operator Xor.

$$T_0(V, v_0') : (v_0' == \overline{v_0})$$
$$T_1(V, v_1') : (v_1' == (v_0 \otimes v_1))$$
$$T_2(V, v_2') : (v_2' == (v_2 \otimes (v_0 \wedge v_1))).$$

## 2.2 Binary Decision Diagrams

A binary decision diagram (BDD) is a DAG (directed acyclic graph) representation of a boolean function. A BDD is composed of two sink nodes and several non-sink nodes. The two sink nodes, labeled 0 and 1, represent the corresponding boolean values. Each non-sink node is labeled with a boolean variable $v$ and has two outgoing edges labeled 1 (or *then*) and 0 (or *else*). Each non-sink node represents the boolean function corresponding to its 1 edge if $v = 1$, or the boolean function corresponding to its 0 edge if $v = 0$.

An ordered binary decision diagram (OBDD) is a BDD with the constraint that the variables are ordered, and every root-to-sink path in the OBDD visits the variables in ascending order.

A reduced ordered binary decision diagram (ROBDD) is an OBDD where each node represents a distinct logic function. This representation is a canonical BDD representation and the most compact representation possible for a given boolean function and a variable ordering.

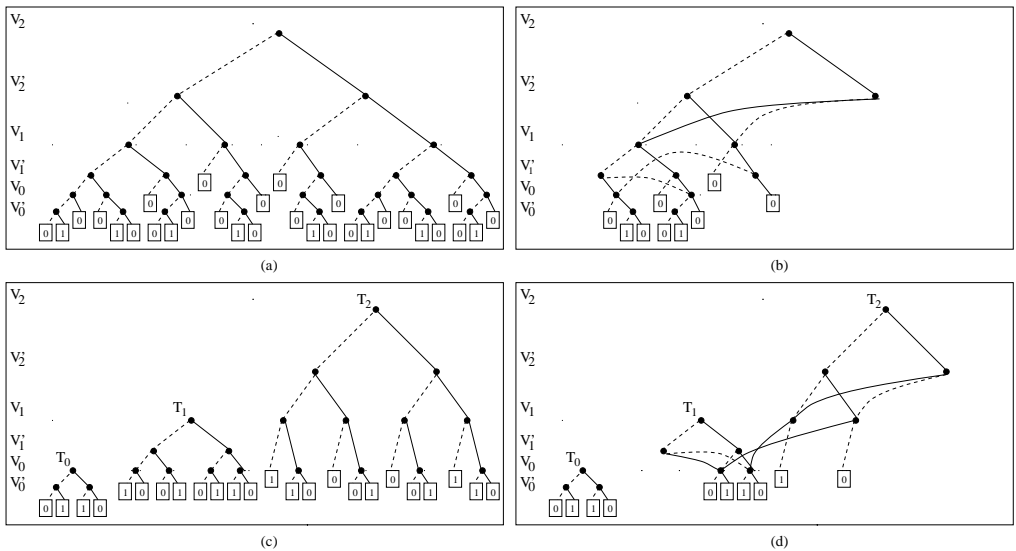

Figure 2: 3-bit counter transition relation (a),(b) and partitioned transition relation (c),(d)

Figure 2 (a),(b) shows the OBDD and ROBDD (respectively) representations of the transition relation function for the 3-bit counter. The dashed lines are the 0 edges and the solid lines are the 1 edges. ROBDDs have only two leaf nodes, one with 1 and one with 0. We drew them several times to enhance readability. ROBDDs can also use complement edges, which produces an even more compact representation. We did not use complement edges, also for reasons of readability. Figure 2 (c),(d) shows the OBDD and ROBDD representations of the partitioned transition relation of the 3-bit counter. The variable order $v_2, v_2', v_1, v_1', v_0, v_0'$ was used in all the representations. Variable ordering algorithms in model checking place the next state variable $v_i'$ adjacent to the present state variable $v_i$.

For the rest of this document we will refer to ROBDDs as BDDs (unless we explicitly state otherwise).

Bollig and Wegener (1996) proved that finding an optimal variable ordering is an NP-complete problem. An order is optimal if it yields a BDD with the smallest number of nodes. Bryant (1986) pointed out that variable ordering greatly influences the size of the BDD. He showed that for a boolean function, one variable ordering may yield a BDD that is exponential in the number of variables, while a different ordering may yield a BDD of polynomial size.

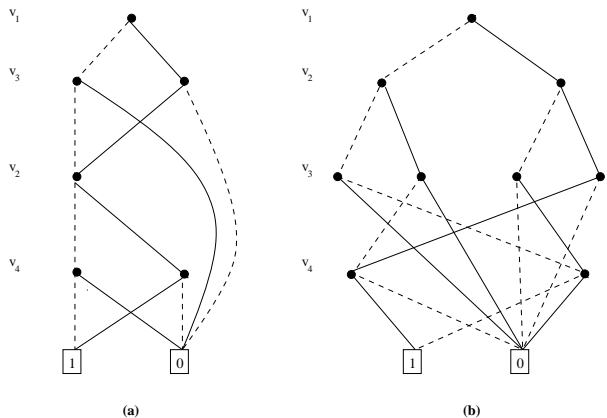

Figure 3: ROBDDs for the function $F(v1, v2, v3, v4) = (v1 = v3) \wedge (v2 = v4)$

Figure 3 gives an example of the effect of variable ordering on the BDD size for the function $F(v1, v2, v3, v4) = (v1 = v3) \wedge (v2 = v4)$. In (a) the variable ordering is $v1, v3, v2, v4$ and in (b) the variable ordering is $v1, v2, v3, v4$.

Various algorithms have been developed for variable ordering. *Exact algorithms* (Ishiura, Sawada, & Yajima, 1991; Drechsler, Drechsler, & Slobodova, 1998; Friedman & Supowit, 1987) are algorithms that find the optimal order. These algorithms use a method similar to dynamic programming with pruning to find the optimal order. Due to the complexity of the problem, exact algorithms are only practical for small cases, and one usually has to turn to other heuristic methods. These heuristic methods can be roughly divided into two groups.

1. *Static Ordering* (Aziz, Tasiran, & Brayton, 1994; Butler, Ross, & Rohit Kapur, 1991; Chung, Hajj, & Patel, 1993; Fujii, Ootomo, & Hori, 1993; Jain, Adams, & Fujita, 1998; Fujita, Fujisawa, & Kawato, 1988; Malik, Wang, Brayton, & Sangiovanni-Vincentelli, 1988; Touati, Savoj, Lin, Brayton, & Sangiovanni-Vincetelli, 1990) which try to find a good ordering before constructing the BDD. Most of these algorithms are based on the topological structure of the verified system.

2. *Dynamic Ordering* (Rudell, 1993; Meinel & Slobodova, 1998; Bollig, Lobbing, & Wegener, 1995; Meinel & Slobodova, 1997; Meinel, Somenzi, & Theobald, 1997; Ishiura et al., 1991; Bern, Meinel, & Slobodova, 1995; Fujita, Kukimoto, & Brayton, 1995; Mercer, Kapur, & Ross, 1992; Zhuang, Benten, & Cheung, 1996; Drechsler, Becker,

& Gockel, 1996; Panda & Somenzi, 1995; Panda, Somenzi, & Plessier, 1994), which given a BDD with some variable order, reorder the variables in the hope of finding a smaller BDD.

In model checking procedures, variable ordering is a central component. At the initial phase of model checking, when the system is translated into a BDD representation, *Static Ordering* is used. The order built at this stage greatly influences the memory usage during the whole computation. However, since model checking keeps producing and eliminating BDDs, the variable order should be changed dynamically in order to effect the size of the current BDDs. *Dynamic Ordering* is used in order to achieve this goal. It is applied by the model checking procedure whenever the size of the BDDs reaches a certain threshold.

Since our work introduces a static ordering algorithm based on machine learning, the next subsection presents a review of the existing static algorithms. Most of these algorithms were developed for combinational circuits (i.e., models whose outputs depend only on their current inputs and not on inputs of previous cycles) and were described with hardware terminology. In order to simplify the description, we will describe them with the terminology we have used so far.

## 2.3 Static Ordering

Static ordering algorithms try to find an initial good order for the BDD. To do so, they extract topological data from the model and use this data to determine an order. All the algorithms convert the model, described by a set of next state functions, into a directed graph known as the *model connectivity graph*. Vertices in the graph are variables and boolean operations (gates). A variable vertex represents a variable, while a gate vertex represents a function. The edges $n_i \rightarrow n_j$ in the graph are between $n_i$, which is either a variable or gate vertex, and $n_j$, which is a gate vertex. An edge $n_i \rightarrow n_j$ is placed if the function represented by $n_i$ is an operand (i.e., an immediate subfunction) of the function represented by $n_j$. We can divide the static algorithms into four groups that differ in the way they use the graph information.

### 2.3.1 Graph Search Algorithms

The method suggested by Malik et al. (1988) assigns to each vertex a *level* metric and orders the variables in decreasing level value. The level of vertices with no out edges is set to be zero and the level of every other vertex $(v_i)$ is set to be $level(v_i) = max_{v_j|v_i \rightarrow v_j}(level(v_j)+1)$. This method resembles a BFS (breadth first search) which originates in nodes that have no out edges, and progresses backwards in the model. Fujita et al. (1988) proposed executing a DFS (depth first search) from the vertices with no out edges, and progressing backwards. Variables in this algorithm are added in post order form.

The algorithms of Malik et al. and Fujita et al. were designed for cases where only one function should be represented in a BDD. This is hardly ever the case in model checking. Butler et al. (1991) adapted the algorithm of Fujita et al. to models with multiple starting points (that is, multiple vertices with no out edges). Their heuristic guides the algorithm to select the first vertex as the vertex that represents the function which depends on the maximum number of variables. This heuristic also guides the search to advance (backwards) from an inner vertex to the vertex that leads to the maximum number of different variables. A tie breaking heuristic (Fujita, Fujisawa, & Matsunaga, 1993) for the enhanced algorithm advises selecting (in case of a tie) the vertex with the maximum number of out edges.

The DFS-based methods *append* the variables to the variable order. Another DFS-based algorithm relies on *interleaving* the variables in the order (Fujii et al., 1993). The algorithm adds a variable after the variable which precedes it in the DFS order.

### 2.3.2 GRAPH EVALUATION ALGORITHMS

Graph evaluation algorithms use the model graph to evaluate the model variables and to perform guided search based on their evaluation values. Minato et al. (1990) propagate values backward through the graph, starting from vertices with no out edges, whose value is set to 1. In vertices of boolean operations, the values on the out edges are summed and the value obtained is divided equally between the in edges. This is done recursively until a vertex of a variable is reached. At variable vertices the propagated values are accumulated as the variable evaluation value. The order is constructed by iteratively adding the variable with the highest value, removing it from the graph, and updating the values.

Chung et al.(1993) proposed two algorithm frameworks. The first framework is composed of two sweeps. In the first sweep each vertex is assigned a value. The values are set by a propagating algorithm that starts from variable vertices with no in edges and advances forward (by their out edges) to all the vertices in the graph. In the second sweep a guided DFS initiated from vertices with no out edges is executed. This search is executed backward in the graph and is guided by the maximal value. This means that the order of traversal among vertex ancestors is according to their assigned value. A number of heuristics to compute the values of the vertices were proposed:

1. *Level-Based* sets the value of variables with no input edges to be zero. The value of the other vertices is set to be the maximal vertex value over its inputs plus one.

2. *Fanout-Based* propagates two values through the graph (one for each boolean value). At a boolean operation vertex the values are not summed and passed along. Rather, they are computed according to the boolean operation at the vertex. The initial values are of variables with no input edges. Their value is set to be the number of out edges the variable has.

In the second framework proposed by Chung et al., the shortest distance between each pair of variables is computed. The total distance of a variable is computed as the sum of its distances to all the variables. The variable with the lowest total distance is selected as the first variable. The next variable is selected as the closest variable to the last ordered variable. Ties are broken according to the distance to previous ordered variables.

All the graph evaluation algorithms try to order the variables so that the variable that most influences the model's next state functions will be first. The algorithms differ by the methodology they use to order the other variables. Some algorithms order them so that variables which substantially influence the model's next-state functions are placed higher in the variable order (toward the beginning of the order). Other algorithms place the other variables according to their proximity to previously ordered variables.

### 2.3.3 Decomposition Algorithms

Decomposition algorithms break down the model into parts. The algorithms then solve two different problems. The first is finding a good order for each part, and the second is finding the order of the parts. The order is constructed by combining the solutions of the two problems.

The algorithm of Malik et al. was extended and adapted for finite-state machines (FSM) by Toutai et al. (1990). In their algorithm, a model is decomposed to its next state functions, each of which is considered separately. Variables of each next state function are ordered according to Malik et al. The next state functions are then ordered by a cost function. They are ordered so that functions with many overlapping variables will be adjacent. The variable order is obtained by adding the variables of the next state functions according to the order of the parts, while removing variables that already exist.

The algorithm of Aziz et al. (1994) decomposes the model in a different way. A model is a hierarchical composition that is constructed by joining a number of internal parts that pass information among themselves. Usually, there is less communication among the parts than within them. Variables of an internal part tend to depend highly on one another. The algorithm uses a *process communication graph (PCG)*, which models the hierarchical structure of the model and the communication between the parts. In a PCG each vertex is an internal part, and an edge $i \rightarrow j$ connects vertex $i$ and vertex $j$ if part $j$ depends on a bit of part $i$. The PCG has parallel edges $i \rightarrow j$, one for each bit value in $i$ that $j$ depends upon. Alternatively, the edges could be weighted.

Given an order of the parts, an upper bound on the BDD size of the model can be computed. The computation is based on the size of the parts and the amount of communication between them. Heuristics guided by the upper bound are applied in order to determine the order of the parts. The order of the variables in each part is decided by one of the previous ordering algorithms.

### 2.3.4 Sample-Based Algorithms

Sample-based static algorithms (Jain et al., 1998) are not real static algorithms in the sense that they do not create the order based on information extracted from the model description. Sample algorithms perform tests on parts of the model (building transition relations and reachable states). For each part, a number of orders are evaluated. The good orders are then merged to create a complete order for the model. Sampling algorithms use "traditional" algorithms in order to find the candidate orders for the parts. These candidate orders are then checked by the sampling algorithm.

### 2.3.5 Summary

A majority of the graph search algorithms and graph evaluation algorithms were developed for other problems and adapted for symbolic model checking. Some of the algorithms were developed in the context of combinational circuits, while others were developed for the simple case of one function. In symbolic model checking the models are rarely combinational (their outputs almost always depend also on inputs of previous cycles), and there is more than one function to display. Adapting the existing algorithms to conform to the needs of symbolic model checking has had various degrees of success. Most of the adapted algorithms are heuristic and apply a simple rule with some logical reasoning behind it.

The decomposition algorithms are either heuristic or provide a theoretical upper bound. However, the bounds they use are rarely realistic; for most models we require much smaller BDDs. The algorithms are also based on decomposing the model into parts and solving the ordering of each part using graph search algorithms. Thus, they also inherit the drawbacks of these algorithms.

Despite the efforts that have been invested and the many algorithms that have been developed for static ordering, the results are not yet satisfactory. The produced BDDs are too large to manipulate, and dynamic ordering must be applied. One problem with the above approaches is their generality: they do not utilize domain-specific knowledge. Domain-specific knowledge is essential for solving the majority of complex problems. It is also difficult to retrieve. In the next subsection we discuss machine learning methods for acquiring domain-specific knowledge for ordering tasks.

## 2.4 Learning to Order Elements

Learning to order elements can be done by first trying to induce a partial order, which can then be used for generating a total order. In this context, a partial order is usually called a *preference predicate*. Preference predicate induction is based on a set of tagged pairs of elements where the binary tag identifies the preferred element. Broos and Branting (1994) present a method for inducing a preference predicate using nearest neighbor classification. The distance between an untagged pair and each tagged pair is computed as the sum of distances between the corresponding elements. The closest tagged pair is selected. The preferred element of the untagged pair is the one matching the preferred element in the tagged pair.

Utgoff and Saxena (1987) represent a pair $A, B$ by the concatenated feature vector $\langle a_1, \ldots a_n, b_1, \ldots b_n \rangle$. The preference predicate is a decision tree induced from these examples.

Utgoff and Clouse (1991) represent a preference predicate by a polynomial. Let $A = \langle a_1, \ldots a_n \rangle$, and $B = \langle b_1, \ldots b_n \rangle$ be a pair of elements represented by feature vectors. Let $w_1, \ldots, w_n$ be a set of weights. The preference predicate $P$ is defined as follows:

$$P(A, B) = \begin{cases} 1 & \sum_{i=1}^{n} w_i(a_i - b_i) \geq 0 \\ 0 & otherwise \end{cases}$$

Each example represents a linear constraint and the weights are found by solving the set of constraints.

Cohen, Schapire and Singer (1999) extended the above mechanism by allowing any preference function $f_i$ instead of $(a_i - b_i)$ in the above expressions. They also present two methods for generating a total order based on the induced preference predicate. Both methods use the preference predicate to construct a graph where the nodes are the elements to be ordered and a directed edge is placed between two elements that have a precedence relation. Two algorithms for inferring the order from the graph are given. The first defines for each node a degree which equals the sum of the outgoing edges minus the sum of the incoming edges. The order is then constructed by selecting the node with the greatest degree and removing its edges from the graph. The second algorithm constructs the order in two stages. In the first stage, all the strongly connected components of the graph are found, and they are ordered according to the dependencies between them. In the second stage the elements of each component are ordered using the first algorithm.

## 3. A Learning Algorithm for Static Variable Ordering

Producing a good variable order requires extensive understanding of BDDs and their relation to the model they represent. Such knowledge can be manually inserted by a human expert. However, this task is too complex for large models. Therefore, it is rarely done. Existing static ordering algorithms use relatively simple heuristic rules that are based on expert knowledge. These rules look at the model structure to compose the ordering. Since the rules are to be applied to all variables in all the models, they are general and thus limited in the ability to produce good orders. Alternatively, we can try to build a program that automatically acquires more specific knowledge based on ordering experience. In this section we present such an algorithm.

The first step in building such a learning algorithm is deciding what knowledge we wish to acquire from the ordering experience. The existing ordering algorithms demonstrate that the precedence relation between variables is a key consideration for the order creation. The graph search algorithms and the search-based graph evaluation algorithms try to place a variable after the variables that influence its next state value. Generally, a variable order of $n$ variables yields $\binom{n}{2}$ *precedence pairs*. A precedence pair $v_i \prec v_j$ denotes that variable $v_i$ should precede $v_j$ in the variable order. For example, the variable order $a, b, c, d$ yields the precedence pairs $a \prec b$, $a \prec c$, $a \prec d$, $b \prec c$, $b \prec d$, $c \prec d$.

The above task of learning precedence pairs can be transformed into a *concept learning task*. A concept learning task is defined by:

- A *universe* $X$ over which the concept is learned;

- A *concept* $C$ – a subset of items in $X$ that we want to learn (usually marked by its associated boolean characteristic function $f_c$);

- A set of *examples* – pairs of the form $\langle x, f_c(x) \rangle$, where $x \in X$;

- A set of *features* – functions above $X$ that allow generalization.

For many learning tasks it is difficult to transform the problem to the format listed above. It is already clear from the discussion above that the general concept we wish to learn is the set of variable pairs in which the first should precede the second in the variable ordering[1] .

More precisely, we define the universe over which the concept is learned as the set of all pairs $\langle (v_i, v_j), M \rangle$, where $(v_i, v_j)$ is an ordered variable pair comprised of $v_i$ and $v_j$, which are variables in the model $M$. Since we expect that some pairs will have no preferred order, we define a ternary instead of a binary concept. The ternary concept has the following classes:

---

1. In practice, we will need only a small subset of the precedence pairs for constructing a total order.

1. $C_+$, the class of all $\langle(v_i, v_j), M\rangle$ for which it is preferable to place $v_i$ prior to $v_j$ in order to get a good initial order.

2. $C_-$, the class of all $\langle(v_i, v_j), M\rangle$ for which it is preferable to place $v_i$ after $v_j$ in order to get a good initial order.

3. $C_?$, the class of all $\langle(v_i, v_j), M\rangle$ for which placing $v_i$ before $v_j$ is just as likely to lead to a good variable order as placing $v_i$ after $v_j$.

In the following subsections we describe the algorithms for learning and using this concept.

## 3.1 Algorithm Framework

We start with the description of the general framework of the learning algorithm. Our goal is to find variable orders that yield BDDs with small number of nodes. Given a *training model*, the algorithm first generates a set of orders of its variables. We define a utility function $u$ over variable orders as following. Each of the orders is used as the initial order for building the BDD representation of the model[2]. This BDD (denoted M-BDD) includes the model's partitioned transition relation and its set of initial states. The *utility $u$* of a generated order is then defined to be reversely proportional to the the number of nodes in the M-BDD constructed with this order.

A subset that consists of all the variable pairs that appear together in some next-state function is selected by the *example extractor* from all the possible variable pairs. We call such pairs *interacting variable pairs*. For example, if $next(x) = y \vee z$ then $(y, z)$ is an interacting variable pair. The *example tagger* tags each of the selected ordered pairs with one of the classes $C_+$, $C_-$, or $C_?$, based on the evaluated orders. The *tagged pairs* are forwarded to the *feature extractor* which, based on the model, computes for each pair its feature vector. The *learner*, which is an ID3 (Quinlan, 1986) decision tree generator, uses the *tagged feature vectors* to create a *pair precedence classifier*.

Several training models are used in this manner to construct different pair precedence classifiers. When solving a new unseen problem, these pair precedence classifiers are used by the *ordering algorithm* to create a variable order.

The learning framework for creating a pair precedence classifier of a training model is given in Figure 4. The complete data flow is displayed in Figure 5. The following subsections describe in greater detail the components of the framework.

## 3.2 The Training Sequence Generator

The goal of the training sequence generator is to produce orders with high variance in quality which is exploited by the tagger (see Subsection 3.4). The simplest strategy for generating such sequences is by producing random orders. This is indeed the strategy we have used in the experiments described in this paper. One potential problem with this approach is with domains where good orders (or bad orders) are rare. In such a case, a random generator will not necessarily produce sequences with the desired diversity in quality.

---

2. We use the SMV (McMillan, 1993) system for this purpose.

Input : Training Model
Output : Precedence Classifier

1. Create sample orders.

2. Use SMV to evaluate the utility of each sample order by the M-BDD size.

3. Find the interacting variable pairs of the training model.

4. Based on the evaluated sampled orders, tag each ordered pair that is based on an interacting variable pair.

5. Transform each tagged pair to a tagged feature vector.

6. Create a classifier based on the tagged feature vectors.

Figure 4: Training model precedence classifier construction

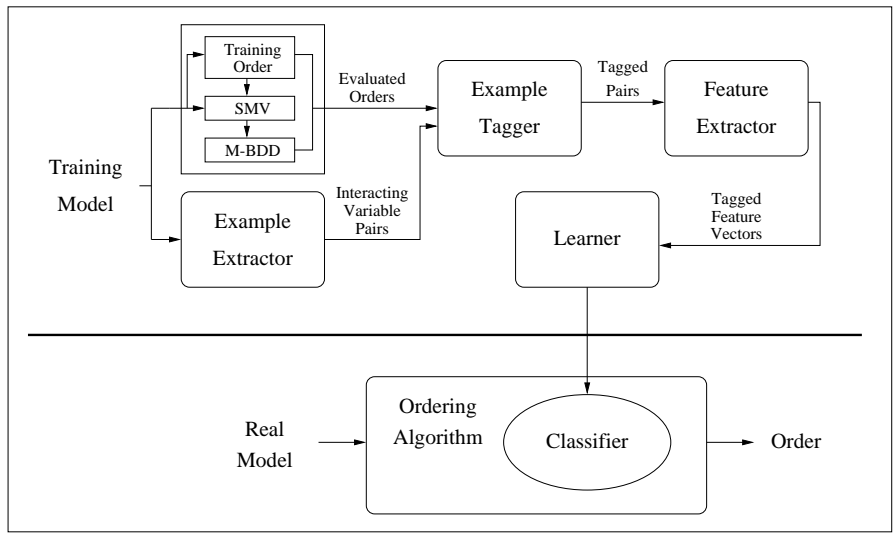

Figure 5: Data flow

An alternative approach is to actively try producing orders that are very good and orders that are very bad, therefore creating a large diversity in quality. One way of producing a good order is by taking the orders that are the result of the dynamic ordering process.

Another option is by using an existing static ordering algorithm. One interesting idea is to try and bootstrap the process by using the results of the adaptive ordering algorithm as training examples thus resulting in progressively more diverse input.

### 3.3 The Example Extractor

Given a set of $n$ variables, we can extract $n * (n-1)$ example ordered pairs for training. But should we actually use all these ordered pairs as examples?

There are two main reasons for being selective about what examples to use:

1. Each example carries computational costs associated with tagging, feature extraction, and the added computation by the induction procedure.

2. Noisy examples are known to have harmful effect on the induction process.

The process of selecting a subset of examples, to be tagged, out of a set of untagged examples is called *selective sampling*. There are two common ways of performing selective sampling. One is by automatic methods that use various general metrics for selecting informative examples (Lindenbaum, Markovitch, & Rusakov, 1999). The other way is by using domain specific heuristics about the potential of an example to be informative.

In this work we use the second approach. Consider a function $f$ over $m$ variables, represented within a BDD of $n$ variables (where $m \leq n$). The number of nodes used to represent $f$ depends only on the relative order of the $m$ variables. This means that changing the order of the other $n - m$ variables would not influence the BDD representation of the function $f$.

The BDD representation of a model to be checked consists of the initial states of the model and the next-state functions of the variables. Since the BDD representation for the initial states is typically small, we do not take it into account. Therefore, when looking for examples, we consider only the next-state functions. Usually, each such function is defined only over a subset of all the model variables. Thus, the order of a pair of variables $(v_i, v_j)$, that do not appear together in any next-state function is less likely to affect the quality of the generated order. We therefore filter out such pairs.

### 3.4 The Example Tagger

An ordered variable pair $(v_i, v_j)$ should be tagged as belonging to $C_+$ if it is preferable to place $v_i$ before $v_j$. Let $V = \{v_1, \ldots, v_n\}$ be the set of variables of a given model. Let $O$ be the set of all possible orderings over $V$. Let $O_{v_i \prec v_j}$ be the set of all $o \in O$ where $v_i$ precedes $v_j$. The ordered variable pair $(v_i, v_j)$ is defined to be *preferable* to $(v_j, v_i)$ if and only if

$$Average\{u(o)|o \in O_{v_j \prec v_i}\} \leq Average\{u(o)|o \in O_{v_i \prec v_j}\}.$$

Since it is not feasible to evaluate all the possible orders, we sample the space of possible orders, evaluate them and partition the samples to two sets as above. As the averages now only estimate the real averages, we replace the term "smaller" in the above definition with "significantly smaller." This is determined by the *unpaired t-test*, which tests the significance (with a given confidence) of the difference between the averages of two samples of two populations.

More precisely, for each variable pair $v_i, v_j$, the set of sampled orders $S \subseteq O$ is partitioned into two subsets $S_{v_i \prec v_j} \subseteq O_{v_i \prec v_j}$ and $S_{v_j \prec v_i} \subseteq O_{v_j \prec v_i}$. An unpaired t-test with a predetermined confidence level is used to check if the averages of the set utilities differ significantly. If they do, the ordered pair corresponding to the set with the smaller average is tagged with + and the other ordered pair is tagged with - (meaning that they belong to $C_+$ and $C_-$, respectively). Otherwise, the average difference is not significant, and both ordered pairs are tagged with ? (meaning that they belong to $C_?$).

A more elaborative approach could use the t-value as a weight on how important a particular order is. These weights could solve conflicts in the ordering process. Such a scheme would require, however, a method to incorporate weights into the induction algorithm. One method is by trying to induce a continues function instead of a ternary function.

## 3.5 The Feature Extractor

If we want to generalize from training models to future unseen models, we cannot represent the pairs by the variable names. Rather, we should use a representation that can be used across models. Most induction algorithms require that the examples be represented by feature vectors.

The process of constructing an appropriate feature set is a crucial part of applying a learning algorithm to a problem. It is a common knowledge engineering process where a domain expert comes up with a set of features that *might* be relevant. It is the role of the induction algorithm, then, to find out what combination of features are relevant to the specific problem.

We have come up with a set of features over variable pairs. These features are extracted from the model connectivity graph. Some of these attributes are inspired by traditional static ordering algorithms. The attributes can be categorized into three groups:

- *Variable attributes* are defined on a single variable and try to capture its characteristics in the model. One example is the *variable-dependence* attribute, which equals the number of variables on which a variable depends. This attribute was inspired by the value used by Butler et al. (1991) to guide the DFS search. A higher value indicates that a larger portion of the model's variables are needed to determine the variable's next-state value. Thus, a higher value may indicate that the variable location should be lower in the order. Another example is the *variable-dependency*, which takes the complementary view of variable-dependence. The attribute equals the number of variables that depend on a given variable. A higher value may indicate that the variable should be placed higher in the variable order.

- *Symmetric pair attributes* are defined on a variable pair $v_i, v_j$. These attributes try to capture the strength of the bond between the two variables, as well as that between this pair and the other variables in the model. For example, *pair-minimal-distance* measures the shortest path between the variables in the model connectivity graph. A shorter path can indicate a stronger bond between the variables. The distance-based ordering framework (Chung et al., 1993) uses a similar feature to order variables. Another example is *pair-mutual-dependency*, which counts the number of variables whose next-state function depends on both $v_i$ and $v_j$.

- *Non-symmetric pair attributes* try to capture the relationship between the two variables. For example, the *pair-dependency-ratio* is the ratio between the variable-dependency values of the two variables. If the ratio is relatively high or low, it may indicate the relative order of the two. *pair-ns-distance* evaluates the influence of one variable on the next state value of the other. It does so by measuring the distance between the variables in the subgraph that represents the next-state function.

The complete list of attributes can be found in Appendix A.

## 3.6 The Induction Algorithm

After the feature extraction phase, our data is represented as a set of tagged feature vectors. This type of representation can be used to produce classifiers by many induction algorithms, including decision trees (Hunt, Marin, & Stone, 1966; Friedman, 1977; Quinlan, 1979; Breiman, Frieman, Olshen, & Stone, 1984), neural networks (Widrow & Hoff, 1960; Parker, 1985; Rumelhart & McClelland, 1986) and nearest neighbor (Cover & Hart, 1967; Duda & Hart, 1973). We have decided to use decision tree classifiers because of their relatively fast learning and fast classification. Fast classification is especially important since we wish to be competitive with other ordering algorithms and the number of variable pairs we need to classify is large.

Decision trees have been researched thoroughly in the last decade, producing many valuable extensions. One such extension enables the decision tree to give not only the classification of items but also to associate with each such classification a confidence estimation. We have used this variant to allow conflict resolution. This will be described in Section 3.7.3.

## 3.7 The Ordering Algorithm

The outcome of the learning process described in the last four subsections is a set of decision trees, one for each training model.

We could also generate one tree based on the union of generated samples. One advantage of the multiple-tree approach is that we expect the examples from the same model to be more consistent, allowing generating compact trees. In contrast, a set of examples coming from different models is likely to be more noisy, yielding a large tree. In addition, the multiple-tree version allows us using a voting scheme during the ordering process, as described below.

Given a *model* $\mathcal{M}$, the algorithm first extracts the *interacting variable pairs*. Each of the classifiers is then applied to the feature vector representations of these pairs. For each classifier, the classifications of all the pairs are gathered to form a precedence table. These tables are then merged into one table. The *order creation algorithm* uses the merged precedence table to construct the model's variable order. The following subsections describe the components in greater detail. Figure 6 shows the data flow in the ordering algorithm.

### 3.7.1 BUILDING THE PRECEDENCE TABLE

To build a precedence table based on a given classifier, the algorithm asks two questions for each interacting variable pair $v_i, v_j$:

1. Should $v_i \prec v_j$ ?

2. Should $v_j \prec v_i$ ?

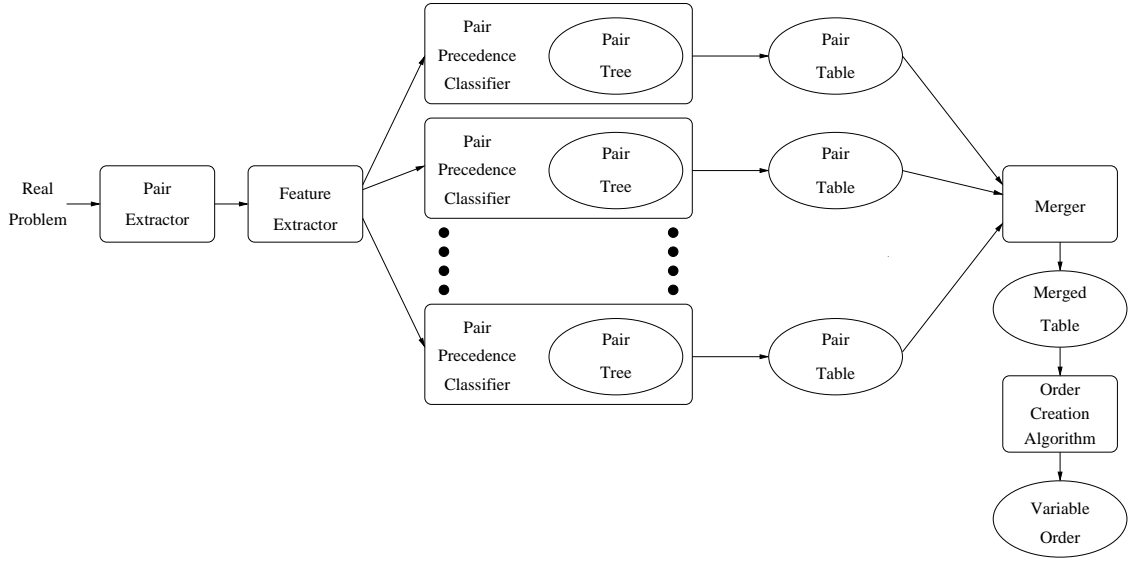

Figure 6: Ordering algorithm data flow

| | $v_i \prec v_j$ ? | $v_j \prec v_i$ ? | $v_i, v_j$ order |
|---|---|---|---|
| 1 | No | No | Unknown |
| 2 | No | Yes | $v_j \prec v_i$ |
| 3 | No | Unknown | Unknown |
| 4 | Yes | No | $v_i \prec v_j$ |
| 5 | Yes | Yes | Unknown |
| 6 | Yes | Unknown | Unknown |
| 7 | Unknown | No | Unknown |
| 8 | Unknown | Yes | Unknown |
| 9 | Unknown | Unknown | Unknown |

Table 2: Pair order table

If the two agree, the pair order is set to the agreed order. If they disagree, the order is set to *unknown*. Table 2 summarizes all the possible answers for the two questions and the resulting pair order.

### 3.7.2 The Merging Algorithm

After constructing the pair precedence tables from the training model's classifiers, we merge the tables using a voting scheme. For each variable pair $v_i, v_j$, we count the number of tables that vote $v_i \prec v_j$ and the number of tables that vote $v_j \prec v_i$. We then decide their pair order according to the majority (ignoring the *unknown* votes).

Assuming that the majority vote chooses the order $v_i \prec v_j$, the confidence for this vote is computed by $\frac{conf(v_i \prec v_j) - conf(v_j \prec v_i)}{vote(v_i \prec v_j) + vote(v_j \prec v_i)}$, where $vote(v_i \prec v_j)$ is the number of tables that vote $v_i \prec v_j$ and $conf(v_i \prec v_j)$ is the sum of the confidence values of these votes. $vote(v_j \prec v_i)$ and $conf(v_j \prec v_i)$ are defined similarly. If this value turns out to be lower than 0.1, we set it to a minimal value of 0.1.

### 3.7.3 CYCLE RESOLUTION

In order to build a total, strict order out of the merged table, the table must not contain any cycle. However, the above algorithm does not guarantee this. We therefore have to apply a cycle resolution algorithm that makes the table cycle-free.

The precedence table can be seen as a directed graph in which the nodes are variables, and there is a weighted edge $v_i \rightarrow v_j$ if and only if $v_i \prec v_j$. There are many possible ways to eliminate cycles in a directed graph. One reasonable bias is removing the least number of edges. This problem is known as the *minimum feedback arc set* and is proven to be NP-hard (Karp, 1972). Approximation algorithms for this problem exist (Even, Naor, Schieber, & Sudan, 1998), but they are too costly for our purposes.

We use instead a simple greedy algorithm to solve the problem. All the constraints (edges) are gathered into a list and sorted in a decreasing order according to their weights (i.e., their confidence). A graph is initialized to hold only the variable vertices. The list of edges is then traversed and each edge is added if it does not close a cycle.

### 3.7.4 PAIR PRECEDENCE ORDERING

At this stage of the algorithm, we hold an acyclic merged precedence table. The last step of the ordering process is to convert the partial order represented by this table to a total order. This is done by topological ordering. At each stage, the algorithm finds all the minimal variables, i.e., variables that are not constrained to follow other unordered variable. From this set, we select a variable $v_{add}$ with maximal fan-out and add it after the last ordered variable. We then add all the variables which are larger than $v_{add}$ but do not appear in any constraint with an unordered variable. We do this because it is desirable to place interacting variables near each other. The *pair precedence ordering (PPO)* algorithm is listed in Figure 7. Figure 8 lists the selection of $v_{add}$ in PPO .

One possible change to the ordering process is to delay the cycle resolution to the last stage. We call this version *cycle resolution on demand*. The modified algorithm does not perform any cycle resolution on the merged table. Instead, the algorithm works with the merged table that may contain cycles. If the table contains a cycle, the algorithm must reach a stage where not all the variables are ordered and there are no minimal variables. In this case the algorithm performs cycle resolution as before and continues the ordering process.

## 3.8 Experiments

We performed an empirical evaluation of the PPO algorithm using models from the ISCAS89 (Brglez, Bryan, & Kozminski, 1989) benchmark. The ISCAS89 benchmark circuits have been used to empirically evaluate many algorithms that deal with various aspects of circuit design (Chamberlain, 1995; Wahba & Borrione, 1995; Nakamura, Takagi, Kimura, & Watanabe, 1998; Long, Iyer, & Abramovici, 1995; Iyer & Abramovici, 1996; Konuk & Larrabee, 1993). We discovered that some of the circuits are insensitive to the initial ordering. This means that the entire sample of initial orders yielded model BDDs of similar sizes. We eliminated these circuits from the set. Out of the remaining circuits we selected those with a number of variables that SMV can handle. We ended up with the following five circuits: s1269 (55), s1423 (91), s1512(86), s4863 (153), s6669 (314). The numbers in parentheses stand for the number of variables in each model.

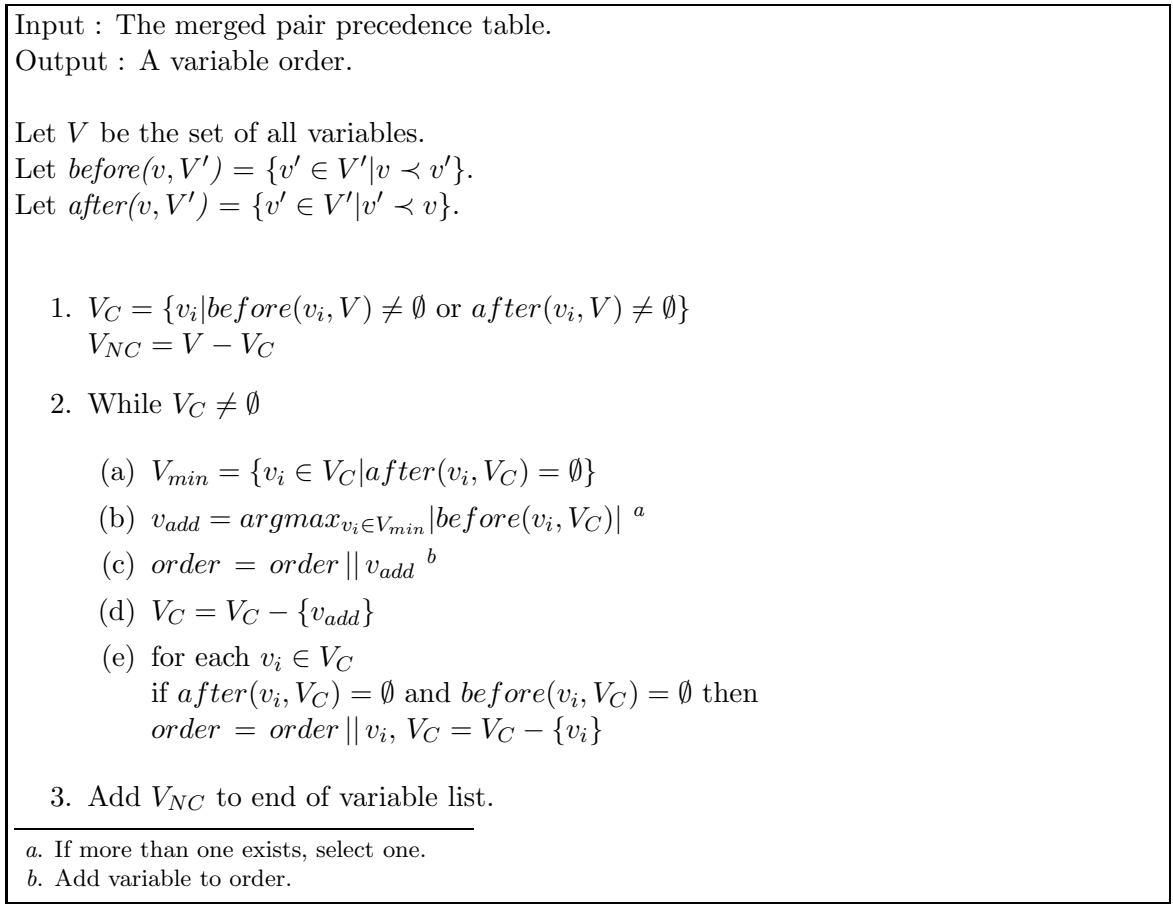

Figure 7: Pair precedence ordering

We began with an offline learning session where the three smaller models (s1269, s1423, s1512) are used as training models. For each of these models we generated 200 random orders and extracted examples as described in the previous section. The algorithm then induced three precedence classifiers in the form of decision trees.

The number 200 was selected since it proved to be sufficiently large. In real application the algorithm can be used as an anytime algorithm where training sequences are generated as long as the user is willing to wait for the offline learner. An alternative approach would keep aside a validation set that would be used for testing the system's performance. The training could have then been stopped when the learning curve flattens.

The algorithm was tested on the two larger models (s4863, s6669). For each of the models, the three learned decision trees were used to generate the merged precedence table.

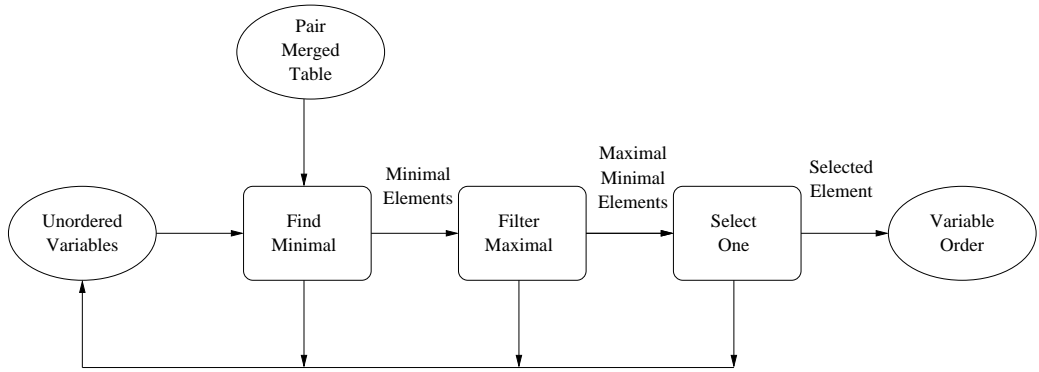

Figure 8: Pair precedence ordering $v_{add}$ selection

Our PPO algorithm (with cycle resolution on demand) was compared to the random algorithm. In addition, we compared our results to two advanced graph search algorithms for static ordering: the DFS *append* algorithm of Fujita et al. (1988) and the *interleave* algorithm of Fujii et al. (1993). In both algorithms we used the adaptation for multiple starting points (Butler et al., 1991) and its expanded version, which includes the tie breaking rule (Fujita et al., 1993). The random results were taken based on 200 variable orders. The two other algorithms were each run 10 times on every model. The performance of the ordering algorithms is measured by the number of nodes in the model BDDs (partitioned transition relation and initial states).

Table 3 and Figure 9 show the results obtained. The table shows that on model s6669, PPO outperformed the random order by more than 300%. On model s4863, PPO outperformed the random order by 5%.

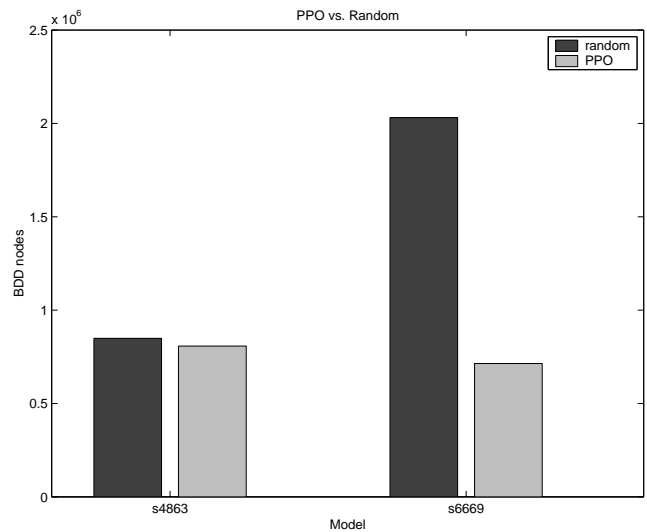

Figure 9: Comparative histogram of PPO vs. Random

| Model | Random | | PPO | |
|-------|---------|-----|---------|-----|
|       | Average | STD | Average | STD |
| s4863 | 849197  | 121376  | 807763 | 100754 |
| s6669 | 2030880 | 1744493 | 713950 | 35446 |

Table 3: Comparative table of PPO vs. Random

The comparison of our algorithm to the two static algorithms is given in Figure 10. The results show that our learning algorithm, after training, becomes competitive with the existing ordering algorithms written by experts.

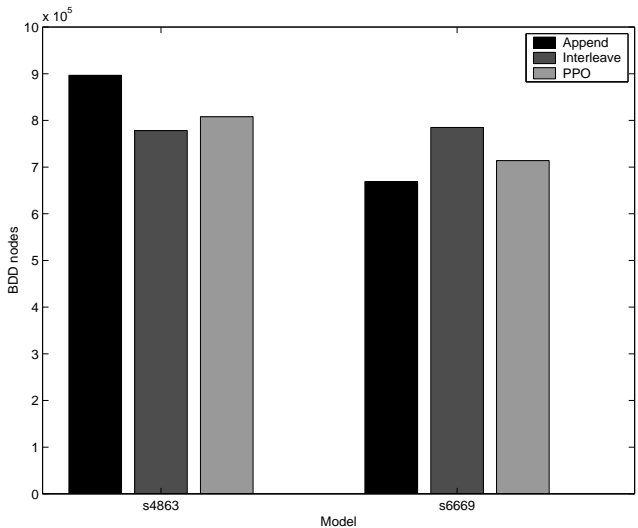

Figure 10: Comparative histogram of PPO vs. Static

To evaluate the utility of the learned knowledge we would like to compare the performance of the ordering process with and without the learned knowledge. Ordering without any learned knowledge is equivalent to random ordering. The comparison of our results to the random-ordering algorithm reveals that the learner indeed induced meaningful knowledge during the learning process. Our method is also much more stable than random ordering on s6669 as indicated by comparing the standard deviation. This large variance in the results of the random ordering is indeed exploited by our tagging procedure as explained in Section 3.4. The small variance in the results obtained by random ordering on s4863 can explain why the improvement obtained by the PPO algorithm is much smaller on this circuit. A more sophisticated training sequence generator, such as those described in Section 3.2, might have been more successful with that circuit.

The comparison to the hand-crafted algorithms may look disappointing at first look since the results of the learning system are not better than the existing algorithms. Recall, however, that we are comparing an automated learning process to human expertise. Most of the works in empirical machine learning make comparisons between the performance of various learning algorithms. It is not common to compare the performance of a learning algorithm with a human expert or an expert system since in most cases it is clear that hand-crafted algorithms would outperform automated learning processes. Since there are hardly any other learning systems that were built to solve the BDD variable ordering problem we could not make the more common comparison between learning systems.

## 4. Learning Context-Based Precedence for Static Ordering

The precedence relation is one of the key considerations used by traditional static ordering algorithms. Another key consideration is the clustering of variables and their subsequent ordering. The algorithms try to place highly interacting variables near each other.

The effect of the variable clustering in a BDD can be seen in the simple example given in Figure 3. In this function, switching the two variables $v_2$ and $v_3$ increases the BDD size by 3 nodes. For this function, all the orders in which the variables of each of the two clusters, $v_1, v_3$ and $v_2, v_4$, are kept together yield the minimal BDD representation. Other variable orders yield a less compact BDD. Thus, in this function, the only key consideration is the compliance with clustering (precedence is not taken into account).

### 4.1 Variable Distance

The above discussion leads to the hypothesis that the distance between variables is an important factor when considering alternative orders. One way to obtain distance information is by learning the distance function between pairs of variables. There are, however, two problems with this approach:

1. The target distance function is not well-defined across models. For example, if we train on small models, the absolute distance function is not likely to be applicable for large models.

2. Information on absolute distances between variables is not sufficient to construct a good ordering. This is because the absolute distance does not uniquely define the order between the variables. In fact, it defines two possible orders, where one is the reverse of the other.

   The example in Figure 11 demonstrates that an order and its reverse can yield BDDs that are significantly different in size. Each of the BDDs in Figure 11 represents two functions, $f_1(a, b, c, d, e) = (a = b = c) \lor (c = d)$ and $f_2(a, b, c, d, e) = (a = b = c) \lor (c = e)$. The absolute distance between the variables in the orders is clearly the same. However, the upper BDD is approximately double the size of the lower one.

   We wanted to check whether in realistic examples reverse orders can yield BDDs that are significantly different in size. We tested models from the ISCAS89 benchmarks and created 5,000 variable orders for each model. For each order, we compared its quality with the quality of the reversed order. We found that in many cases one order was exceptionally good while the reversed one was exceptionally bad. Thus, learning the absolute distance is not sufficient, and more information is needed.

We conclude that there are problems inherent both in learning and in utilizing absolute distances. Still, clustering is a key consideration and should be pursued. We suggest, alternatively, learning the *relative* distance that determines, for variables $v_i, v_j$, and $v_k$, which of $v_j, v_k$ should be closer to $v_i$, given that $v_i$ precedes the other two.

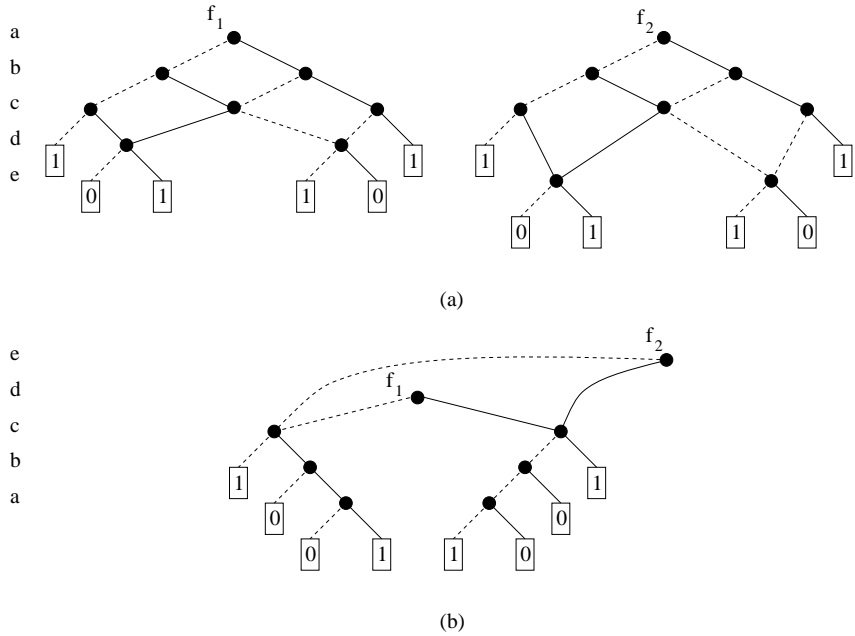

Figure 11: ROBDDs for the functions $f_1(a, b, c, d, e) = (a = b = c) \vee (c = d)$ and $f_2(a, b, c, d, e) = (a = b = c) \vee (c = e)$

The remainder of this section describes a method for learning and utilizing *context-based precedence* to infer the relative distance between variables.

## 4.2 Context-Based Precedence

A *context precedence relation* is a triplet $v_i \prec v_j \prec v_k$: given that $v_i$ precedes $v_j$ and $v_k$, the variable $v_j$ should come before the variable $v_k$. Thus, the context precedence relation adds context to pair ordering decisions.

As in pair precedence learning, we define the universe to be the set of pairs $\langle (v_i, v_j, v_k), M \rangle$ where $v_i, v_j, v_k$ are variables in the model $M$. The universe is divided into three classes, $C_+, C_-, C_?$, as before. Examples for these classes are drawn in the same way. The pair precedence framework can be applied with minor changes to work with context precedence relations. These minor changes are described below.

## 4.3 The Example Tagger

A variable triplet $(v_i, v_j, v_k)$ should be tagged as $C_+$ if, given that $v_i$ precedes $v_j$ and $v_k$, it is preferable to place $v_j$ before $v_k$ (i.e., $v_i \prec v_j \prec v_k$). As in pair precedence learning, we use a set of evaluated variable orders for the tagging. Any set of such orders can be partitioned to three subsets, depending on which of the three variables is first. Given a partition defined by $v_i$ (for example), we can test the order of $v_j$ and $v_k$ using *t-test*, as described in Section 3.4. To reduce the number of noisy examples, we use only the partition that yields the most significant t-test results.

## 4.4 The Feature Extractor

The attributes of a triplet $(v_i, v_j, v_k)$ are computed based on the attributes of the two pairs $v_i \prec v_j$ and $v_i \prec v_k$. Each attribute value is the division/subtraction of two corresponding attribute values from the two pair attributes.

More precisely, assume that the pair $v_i \prec v_j$ has attributes $f_1(v_i, v_j), \ldots, f_n(v_i, v_j)$ and the pair $v_i \prec v_k$ has attributes $f_1(v_i, v_k), \ldots, f_n(v_i, v_k)$. Then the triple $(v_i, v_j, v_k)$ has attributes $f_1(v_i, v_j)/f_1(v_i, v_k), \ldots, f_n(v_i, v_j)/f_n(v_i, v_k)$. If some of the $f_l(v_i, v_k)$ can be 0 then the corresponding attributes are subtracted instead of divided.

As an example consider an attribute $f_l$ which is *pair minimal distance* (see Section 3.5). If $f_l(v_i, v_j)/f_l(v_i, v_k)$ is greater than 1 than the shortest path between $v_i$ and $v_j$ is larger than the shortest path between $v_i$ and $v_k$. This attribute can indicate that $v_k$ should appear closer to $v_i$.

Similarly, if $f_l$ is *pair mutual dependency* then $f_l(v_i, v_j)/f_l(v_i, v_k) > 1$ indicates that the number of variables whose next-state function depends on both $v_i$ and $v_j$ is greater than those depending on both $v_i$ and $v_k$. This may indicate that it is preferable to keep $v_i$ and $v_j$ close together.

## 4.5 The Ordering Algorithm

The outcome of the learning phase is a set of decision trees, one for each model. This is the same as in the case of context-free pairs. In this subsection we describe ways to use these trees for ordering.

### 4.5.1 BUILDING THE CONTEXT PRECEDENCE TABLE

While in the case of pair precedence we had a table of size $n^2$ (where $n$ is the number of variables), we now produce one such table for each context variable. For each table we perform inconsistency elimination similar to that described in Section 3.7.1. Here, however, when we ask the classifier the two questions $v_j, v_k$ and $v_k, v_j$, we add the context variable $v_i$ to the query.

### 4.5.2 PAIR PRECEDENCE ORDERING WITH CONTEXT PRECEDENCE FILTERING

The ordering algorithm uses the pair precedence table in the same way as the PPO algorithm. However, it was often found to be the case that the PPO algorithm had several minimal variables, even after employing the maximal fanout filter. We use the context-based precedence table to further reduce the size of the set of minimal elements. We use the variables in the already ordered sequence as context variables and look at their associated tables. If the set of minimal elements contains a pair of variables constrained as $v_j \prec v_k$ in one of the tables, we eliminate $v_k$ from the set. Figure 12 lists the code which when added to the PPO algorithm, accepts a variable set $V_{add}$ (from which we previously selected randomly), and returns one variable. We call the new algorithm PPO$^{\text{CPF}}$. Figure 13 lists the selection of $v_{add}$ in PPO$^{\text{CPF}}$.

Input : The set of candidate variables to be added, $V_{add}$, and the merged context precedence table.

Output : A variable to be added.

Let $after(v, V_i, V_j) = \{\langle v_i, v_j \rangle \, v_i \in V_i, v_j \in V_j | v_i \prec v_j \prec v\}$.

b.1 $V'_{add} = \{v_i \in V_{add} | after(v_i, V_{InOrder}, V_{add}) = \emptyset\}$

b.2 If $V'_{add} \neq \emptyset$ then
    select randomly one variable from $V'_{add}$
    else select randomly one variable from $V_{add}$

Figure 12: Pair precedence ordering with context precedence filtering

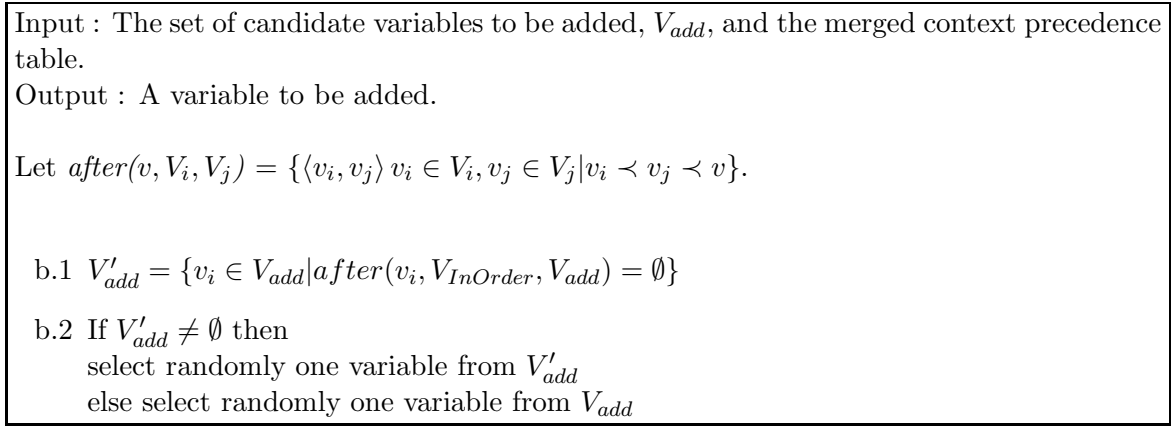

Figure 13: Pair precedence ordering with context precedence filtering $v_{add}$ selection

## 4.6 Experiments

We have evaluated the performance of $\mathrm{PPO}^{\mathrm{CPF}}$, performing off-line learning on the training models followed by ordering of the test models. The results are shown in Figure 14. For comparison we also show the performance of the PPO algorithm and the two expert algorithms.

The $PPO^{CPF}$ algorithm outperforms all the other algorithms on the two tested models. The results show that the context-based precedence relations add valuable information.

We have tested the effect of the resources invested in the learning phase on the performance of the algorithms. Since the learning examples are tagged based on evaluated training orders, and since the evaluation of the training orders is the most resource-consuming operation, we used the number of these orders as the resource estimator. Figure 15 shows the learning curves of our algorithms, that is, it shows how the system performance changes according to the offline resources consumed (the number of training orders evaluated).

Without testing any random order, our system has no knowledge on which to build the precedence classifiers, and thus its performance is equivalent to random ordering. The tagging based on 20 orders is too noisy. While it improves the performance of s6669, it degrades the performance of s4863. Forty orders are sufficient to generate stable tagging, which yields improved classifiers and therefore improved ordering quality.

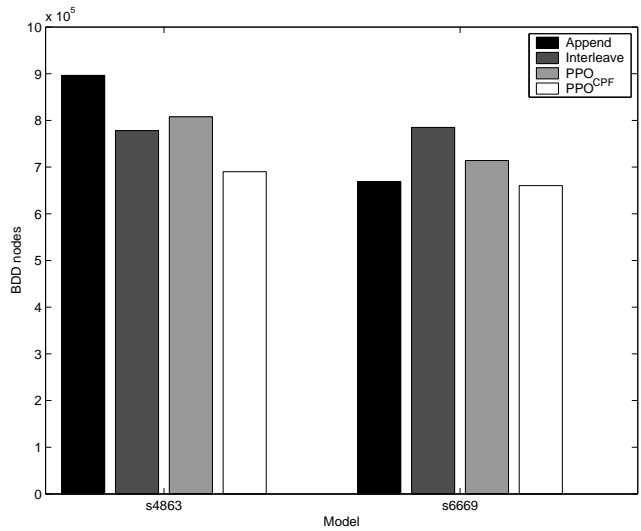

Figure 14: Comparative histogram of ordering algorithms

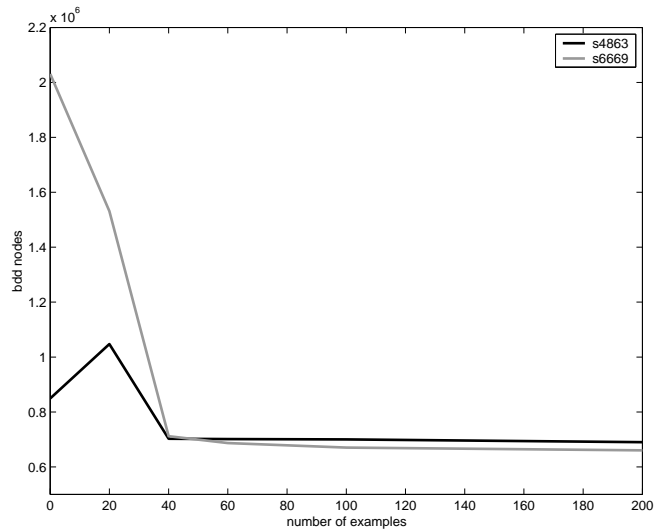

Figure 15: Learning curves of the $PPO^{CPF}$ algorithm for the two testing models

## 5. Discussion

The work described in this paper presents a general framework for using machine learning methods to solve the static variable ordering problem. Our method assumes the availability of training models. For each training model, the learning algorithm generates a set of random orders and evaluates them by building their associated BDDs. Each ordered pair of interacting variables is then tagged as a good example if it appears more frequently in highly valued orders. The ordered pairs are converted to feature-based representations and are then given, with their associated tags, to an induction algorithm. When ordering variables of a new unseen model, the resulting classifiers (one for each model) are used to determine the ordering of variable pairs. We also present an extension of this method that learns context-based ordering.

Our algorithm was empirically tested on real models. Its performance was significantly better than random ordering, meaning that the algorithm was able to acquire useful ordering knowledge. Our results were slightly better than existing static ordering algorithms hand-crafted by experts. This result is significant if we compare it to applications of learning systems to other domains. We would surely appreciate an induction algorithm that produces a classifier with performance comparable to that of an expert system built by a medical expert. A chess learning program that is able to learn an evaluation function that is equivalent in power to a function produced by an expert will be similarly appreciated. We therefore claim that the ability of or learning algorithm to achieve results that are as good as manually designed algorithms indicates strong learning capabilities.

In most learning algorithms, we expect to get better performance when the testing problems are similar to the training problems. In the verification domain, we expect to get good results when the testing and training models come from a family of similar models. There are several occasions in which models are similar enough to be considered a family: Models of different versions of a design under development; models which are reduced versions of a design, each with respect to a different property; models of designs with a similar functionality like ALUs, arbiters, pipelines and bus controllers. Unfortunately, due to the difficulty in obtaining suitable real models for our experiments, we ended up experimenting with training and testing models that are not related. We expect to achieve much better results for related models.

Compared with previous work in machine learning, our precedence relations most resemble these of Utgoff and Saxena (1987). Our ordering approach, in which we construct a total order of elements by finding the precedence relation between them, is in essence the same as that of Cohen, Schapire and Singer (1999). Specifically, the second ordering algorithm of Cohen, Schapire and Singer also uses the topological ordering approach to create an order. Their algorithm initially finds in the precedence graph the connected components and, after ordering them (using topological ordering), finds the order in each connected component. However, since the quality of the final order is determined by the sum of constraints adhered to, all topological orders have theoretically the same quality. We found that in the BDD variable ordering problem not all topological orders have the same quality. Thus, we developed a topological ordering that takes into consideration those features that we recognized as true for variable orders in BDDs.

Our work also differs from previous research in that it introduces the notion of context-based precedence. Using this concept we were able to create an ordering algorithm that produces the best results.

There are several directions for extending the work described here. One problem with our current empirical evaluation is the small number of models. In spite of our extensive search efforts we were not able to find a large set of suitable examples. The majority of the known examples are very simple (compared with real industry problems), producing small model BDD representations with very little variance. We are currently in the process of approaching companies that use model checking. In this way we hope to obtain additional real models, preferably from families of the designs described above.

The attributes of the variable pairs were partially based on substantive research in the field of static algorithms. We could not find such information on which to base context-based variable attributes. Thus, we also based these attributes on those of the variable pairs. Nevertheless, we believe that human experts in this field may have information that can lead to the development of better attributes. The development of such attributes should help to capture in a better way the context-based precedence concept.

Given our current results, an immediate question is whether the concept of precedence pairs (context and non-context) can be extended to triplets, quadruples, etc. Such precedence relations take into account a larger part of the model and thus may possess valuable information. Such an extension, however, could carry high cost during learning and, even worse, during ordering.

Our framework for solving the static variable ordering problem was shown to be valuable in model checking. Model checking is only one field of verification in which BDDs are used. BDDs are also used in verification for simulation and equivalence checking. Our algorithm can be applied for these problems as well. We are unaware of special static variable ordering algorithms for these fields, but if such do exist, variable attributes based on these algorithms should be added.

The most interesting future direction is the generalization of our framework for other ordering problems. Ordering a set of objects is a very common sub-task in problem solving. The most common approach for tackling such a problem is to evaluate each object using a utility function and order the objects according to their utilities. Such an approach is taken, for example, by most heuristic search algorithms. In many problems, however, it is much easier to determine the relative order of two objects than to give each object a global utility value. Few works have applied learning to ordering techniques that are not utility based (Cohen et al., 1999). The algorithms described in Section 3 and Section 4 can be applied to any ordering problem if a method for evaluating training orders is available, and a set of meaningful pair features can be defined.

We believe that the research presented in this paper contributes both to the field of machine learning and to the field of formal verification. For machine learning, it presents a new methodology for learning to order elements. This methodology can be applied to various kinds of ordering problems. For formal verification, it presents new learning-based techniques for variable ordering. Finding good variable ordering techniques is one of the key problems in this field.

## Appendix A. Variable Pair Attributes

The following definitions and symbols will be used in the attribute description:

- $NS(v_i)$ for the next state function of variable $v_i$

- $v_i \triangleright v_j$ to indicate that variable $v_i$ *depends on* variable $v_j$'s value ($v_j \in \mathrm{NS}(v_i)$)

- $v_i \bowtie v_j$ to indicate that variable $v_i$ *interacts with* variable $v_j$ ($v_i \rhd v_j$ and/or $v_j \rhd v_i$)

- *# variables* for the number of variables in the model

## A.1 Variable Attributes

The attributes computed for $v_i$ are

1. *Variable-dependence*: the number of variables upon which $v_i$ depends ($|\{v_j|v_i \rhd v_j\}|$)

2. *Variable-dependency*: the number of variables that depend on $v_i$ ($|\{v_j|v_j \rhd v_i\}|$)

3. *Variable-dependency-size*: the sum of function sizes that depend on $v_i$ ($\sum_{v_j \rhd v_i} |\{v_k \in NS(v_j)\}|$)

4. *Variable-dependency-average-size*: the average function size dependent on $v_i$
   $$\left( \frac{\sum_{v_j \rhd v_i} |\{v_k \in NS(v_j)\}|}{|\{v_j|v_j \rhd v_i\}|} \right)$$

5. *Variable-dependence-dependency-ratio*: the proportion between the number of variables on which $v_i$ depends and the number of variables that depend on it $\left( \frac{|\{v_j|v_i \rhd v_j\}|}{|\{v_j|v_j \rhd v_i\}|} \right)$

6. *Variable-interaction*: the number of variables interacting with $v_i$ ($|\{v_j|v_i \bowtie v_j\}|$)

7. *Variable-dependence-percentage*: the percentage of model variables on which $v_i$ depends $\left( \frac{|\{v_j|v_i \rhd v_j\}|}{\#variables} \right)$

8. *Variable-dependency-percentage*: the percentage of model variables that depend on $v_i$
   $\left( \frac{|\{v_j|v_j \rhd v_i\}|}{\#variables} \right)$

9. *Variable-interaction-percentage*: the percentage of model variables interacting with $v_i$
   $\left( \frac{|\{v_j|v_i \bowtie v_j\}|}{\#variables} \right)$

## A.2 Variable Pair Attributes

The attributes computed for $\langle v_i, v_j \rangle$ are

- Symmetric attributes

  1. *Pair-minimal-distance*: the minimal distance between $v_i, v_j$ in the model graph

  2. *Pair-minimal-distance-eval*: the minimal distance between $v_i, v_j$ in the model graph divided by the number of times it appears

  3. *Pair-minimal-dependency*: the number of variables that depend on the pair with the minimal distance

  4. *Pair-minimal-dependency-eval*: the minimal distance between $v_i, v_j$ in the model graph divided by number of variables that depend on the minimal distance

5. *Pair-minimal-connection-class*: the minimal distance between the $v_i, v_j$ connection class (the operators that can be applied on two variables were divided into classes and the operator that connected the two variables in the minimal distance class was extracted)

6. *Pair-minimal-maximal*: the maximal sized $NS(v_k)$ connecting the pair in minimal distance

7. *Pair-minimal-maximal-eval*: the minimal distance between $v_i, v_j$ in the model graph divided by maximal sized $NS(v_k)$ connecting the pair in minimal distance

8. *Pair-sum-distance*: the sum of distances between $v_i, v_j$ in the model graph

9. *Pair-dependency-ns-size*: the sum of $NS(v_k)$ sizes that are dependent on $v_i$ and $v_j$ $(\sum_{v_k \rhd v_i \,\&\, v_k \rhd v_j} |v_l \in NS(v_k)|)$

10. *Pair-sum-distance-dependency-ratio*: the sum of distances between $v_i, v_j$ in the model graph divided by sum of $NS(v_k)$ sizes that are dependent on $v_i$ and $v_j$

11. *Pair-mutual-dependence*: the number of variables on which both $v_i, v_j$ depend $(|\{v_k | v_i \rhd v_k \,\&\, v_j \rhd v_k\}|)$

12. *Pair-mutual-dependency*: the number of variables that depend on $v_i$ and $v_j$ $(|\{v_k | v_k \rhd v_i \,\&\, v_k \rhd v_j\}|)$

13. *Pair-mutual-interaction*: the number of variables that interact with $v_i$ and $v_j$ $(|\{v_k | v_i \bowtie v_k \,\&\, v_i \bowtie v_k\}|)$

14. *Pair-mutual-ns-dependency*: $v_i$ depends on $v_j$ and $v_j$ depends on $v_i$ - $(v_i \rhd v_j \,\&\, v_j \rhd v_i)$

- Non-Symmetric attributes ( those computed for the pair $\langle v_i, v_j \rangle$ with relevance to $v_i$)

  1. *Pair-ns-distance*: the distance between $v_i, v_j$ in $NS(v_i)$

  2. *Pair-dependence-ratio*: the ratio between the number of variables that $v_i$ depends on and the number of variables that $v_j$ depends on $\left( \frac{|\{v_l | v_i \rhd v_l\}|}{|\{v_m | v_j \rhd v_m\}|} \right)$

  3. *Pair-dependency-ratio*: the ratio between the number of variables that depend on $v_i$ and the number of variable that depend on $v_j$ $\left( \frac{|\{v_l | v_l \rhd v_i\}|}{|\{v_m | v_m \rhd v_j\}|} \right)$

  4. *Pair-interaction-ratio*: the ratio between the number of variables that interact with $v_i$ and the number of variables that interact with $v_j$ $\left( \frac{|\{v_l | v_i \bowtie v_l\}|}{|\{v_m | v_j \bowtie v_m\}|} \right)$

  5. *Pair-dependence-flag*: the number of variables that $v_i$ depends on compared to the number of variables that $v_j$ depends on $\left( \frac{|\{v_l | v_i \rhd v_l\}|}{|\{v_m | v_j \rhd v_m\}|} >= 1.0 \right)$

  6. *Pair-interaction-flag*: the number of variables that interact with $v_i$ compared to the number of variables that $v_j$ interacts with $\left( \frac{|\{v_l | v_i \bowtie v_l\}|}{|\{v_m | v_j \bowtie v_m\}|} >= 1.0 \right)$